\documentclass[10pt,twocolumn,letterpaper]{article}
\usepackage[table]{xcolor}
\usepackage{iccv}
\usepackage{amssymb}
\usepackage{array}
\usepackage{enumitem}
\usepackage{graphicx}
\usepackage{epsfig} 
\usepackage[cmex10]{amsmath}
\usepackage{mathptmx} 
\usepackage{amsfonts}
\usepackage[vlined, ruled ]{algorithm2e}
\usepackage{mdwmath}
\usepackage{hhline}
\usepackage{booktabs}
\usepackage{caption}
\usepackage[caption=false,font=footnotesize]{subfig}
\usepackage{url}
\usepackage{float}
\usepackage[numbers,sort&compress]{natbib}
\usepackage{balance}
\usepackage[english]{babel}
\usepackage[font=footnotesize]{caption}
\usepackage{comment}
\definecolor{Gray}{gray}{0.85}

\newcommand{\eqnref}[1]{Equation~(\ref{eqn:#1})}
\newcommand{\figref}[1]{Fig.~\ref{fig:#1}}
\newcommand{\tblref}[1]{Table~\ref{tbl:#1}}
\newcommand{\secref}[1]{Section~\ref{sec:#1}}
\newcommand{\algref}[1]{Algorithm.~\ref{alg:#1}}

\newcommand{\figbs}{
\begin{figure}[t]
  \centering
  \includegraphics[width=0.4\textwidth,  trim=0in 0.1in 0in 0.1in,
  clip=true]{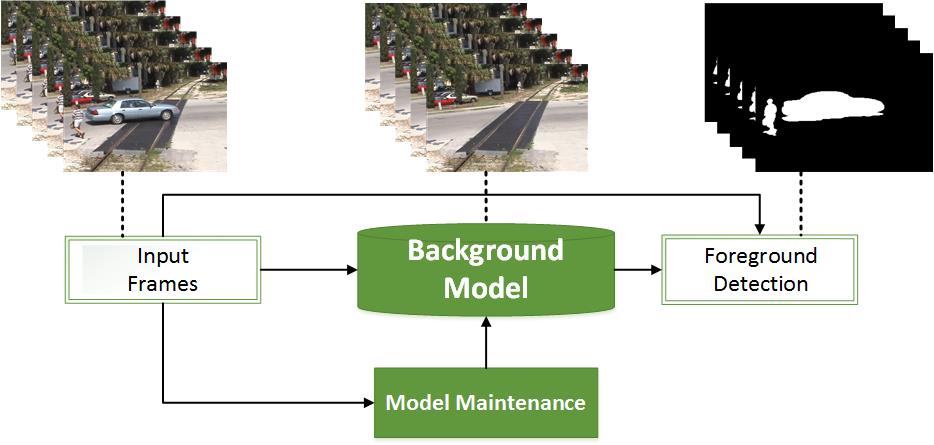}
  \caption{Standard structure common in the majority of the robust background subtraction methods.}
\label{fig:bs}
\vspace{-.2in}
\end{figure}
}

\newcommand{\figbmc}{
\begin{figure*}[t]
 \centering
 \subfloat[]{\label{fig:rpca}\includegraphics[width=0.33\textwidth, trim=3in 0in 3in 0in,
  clip=true]{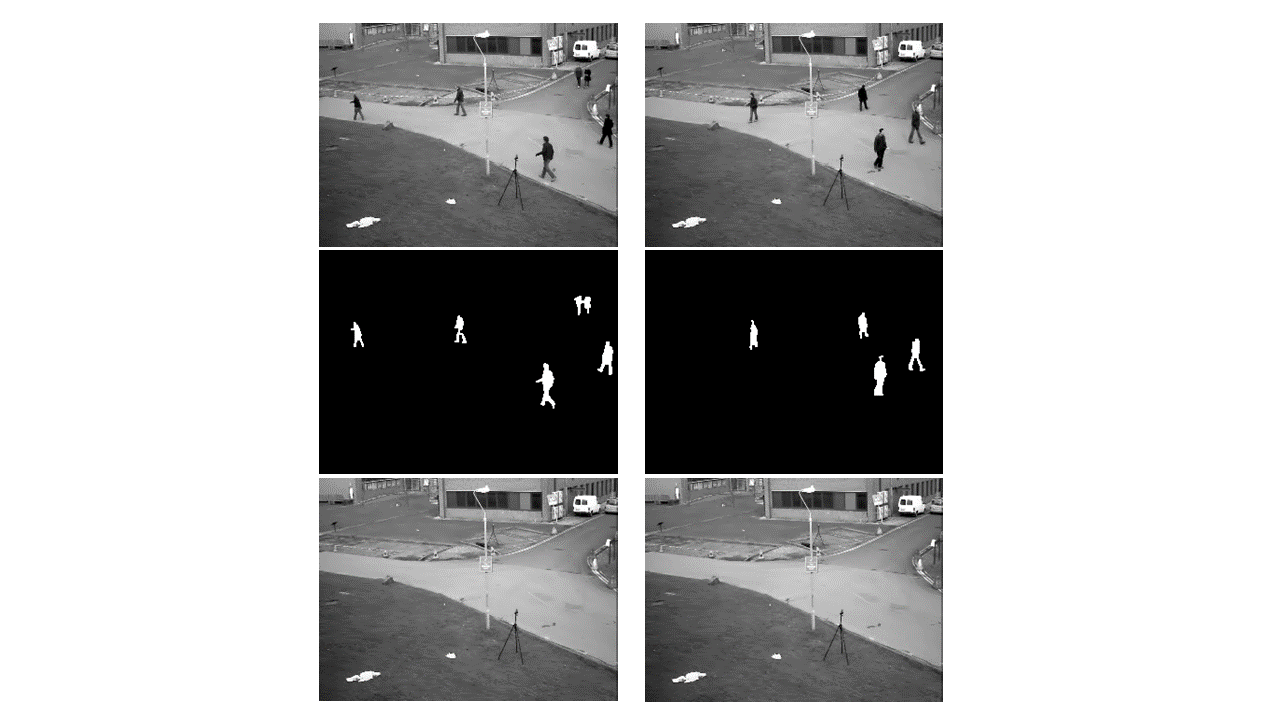}}
 \subfloat[]{\label{fig:rmc}\includegraphics[width=0.33\textwidth,trim=3in 0in 3in 0.25in,
  clip=true]{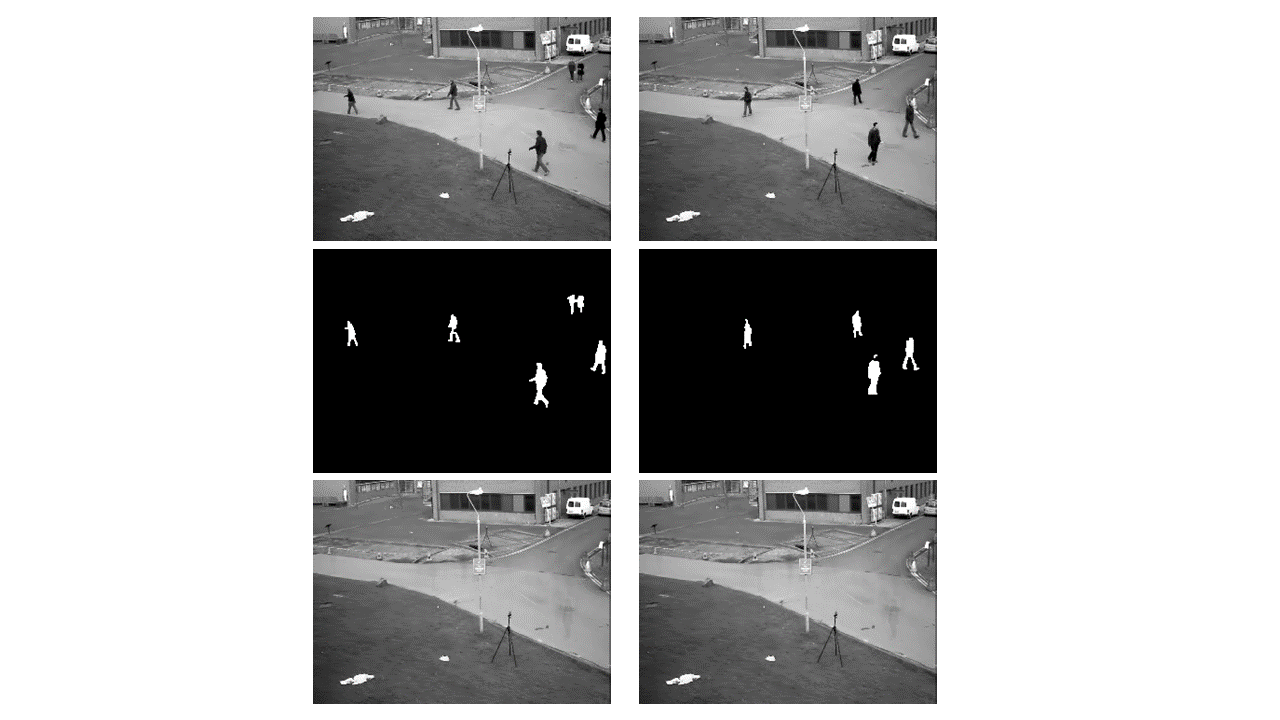}}
  \subfloat[]{\label{fig:frmc}\includegraphics[width=0.33\textwidth,  trim=3in 0in 3in 0in,
  clip=true]{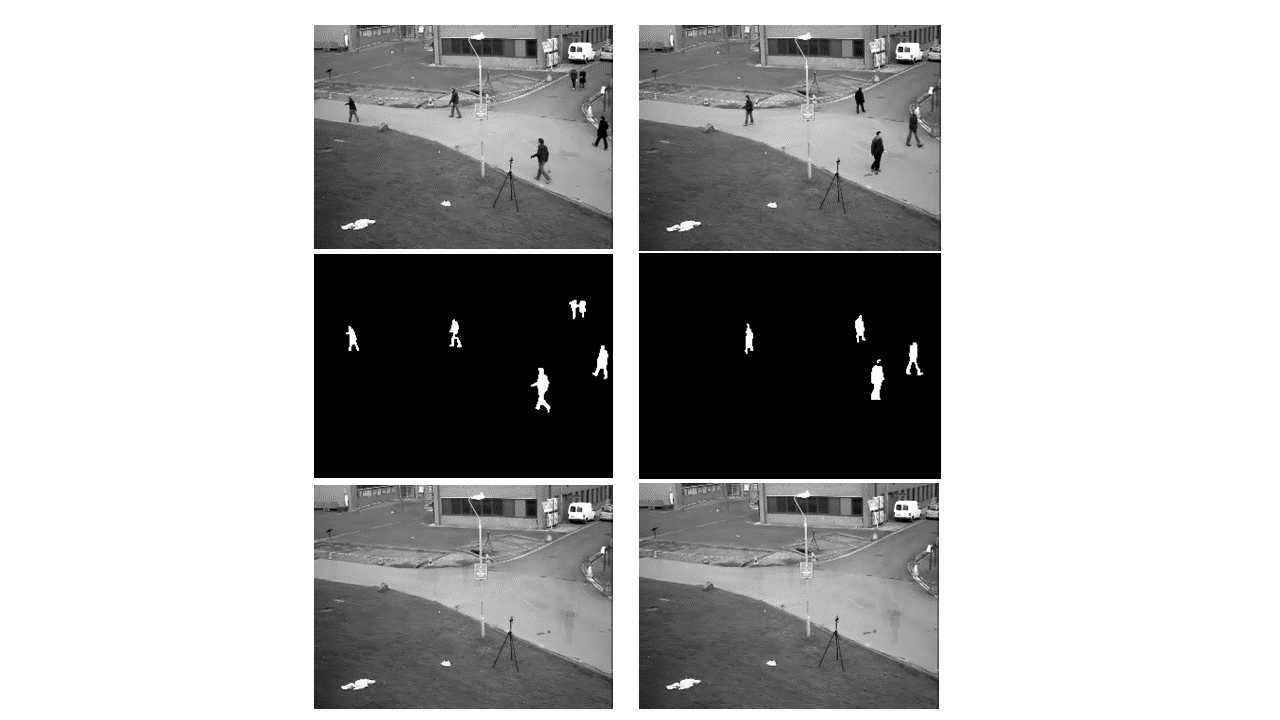}} 
  \caption{Result of applying the three methods (a) RPCA, (b) RMC, and (c) fRMC on BMC wandering students video demonstrated for two different frames; first row is the original frame, second row is the binary mask of the foreground by thresholding the detected foreground, and last row is the background image.}
\label{fig:bmc}
\vspace{-.2in}
\end{figure*}
}

\newcommand{\figsabs}{
\begin{figure*}[t]
 \centering
 \subfloat[]{\label{fig:rpca}\includegraphics[width=0.33\textwidth, trim=3in 0in 3in 0in,
  clip=true]{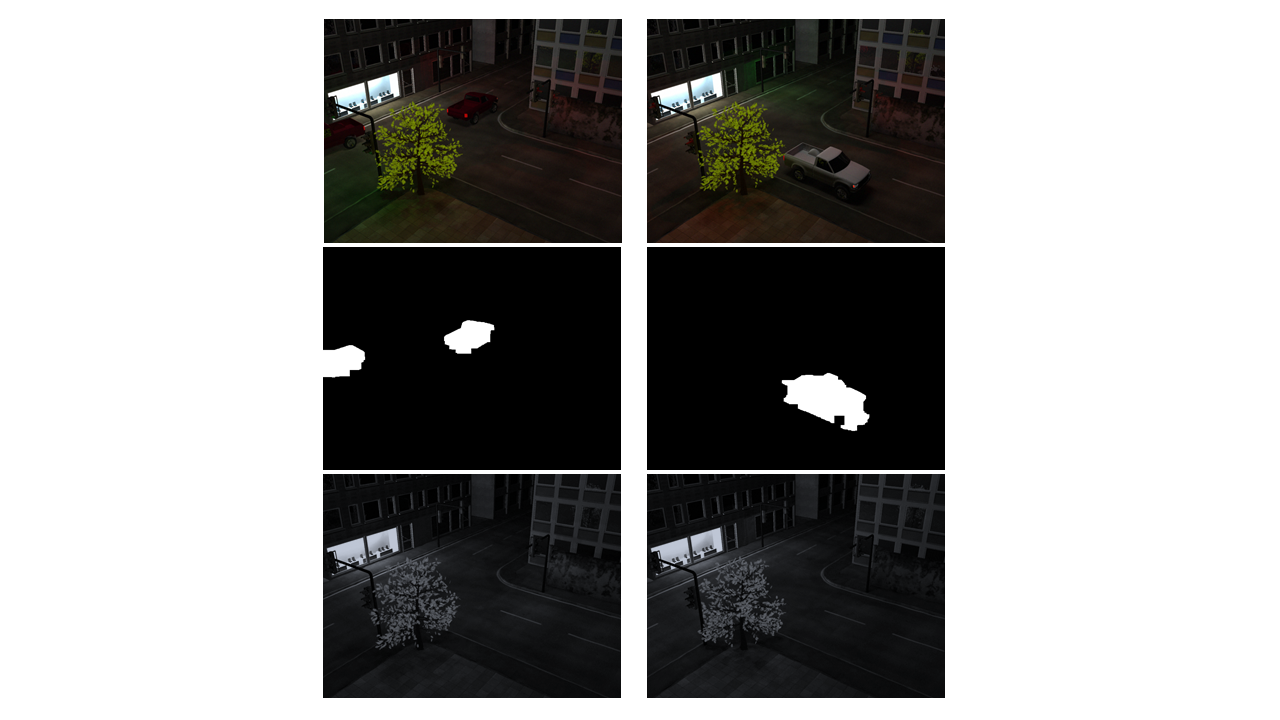}}
 \subfloat[]{\label{fig:rmc}\includegraphics[width=0.33\textwidth,trim=3in 0in 3in 0.25in,
  clip=true]{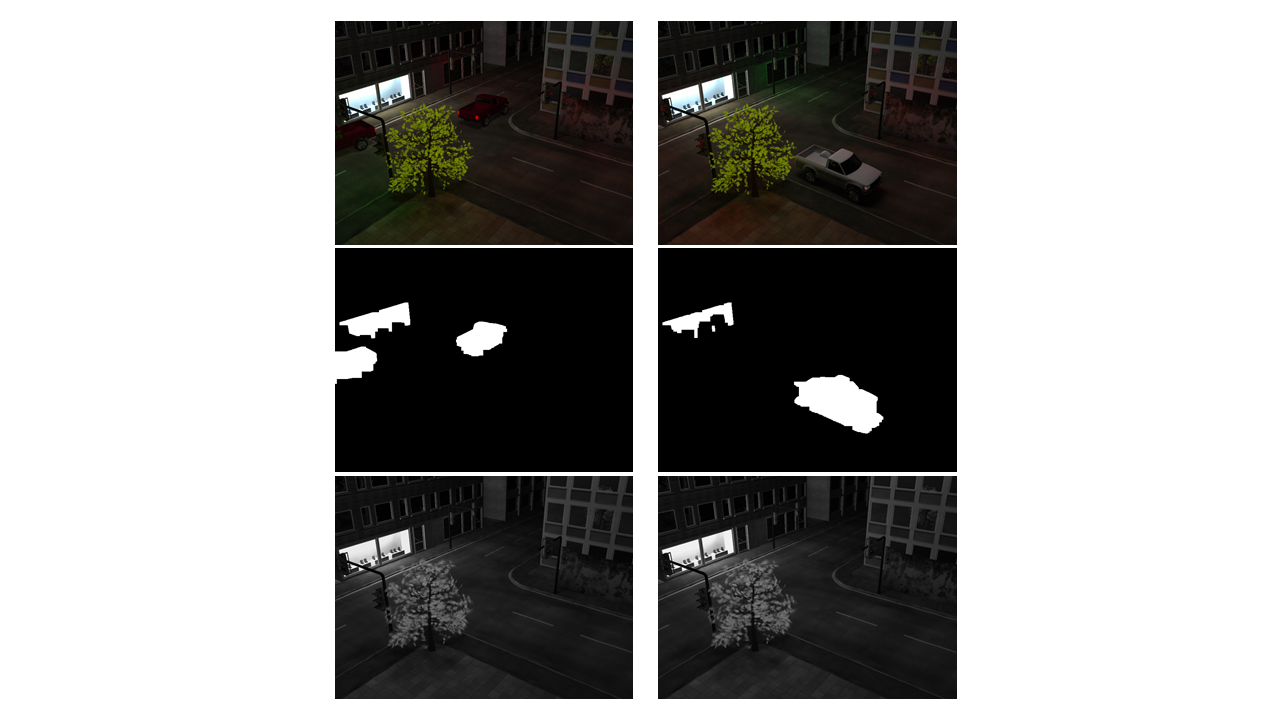}}
  \subfloat[]{\label{fig:frmc}\includegraphics[width=0.33\textwidth,  trim=3in 0in 3in 0in,
  clip=true]{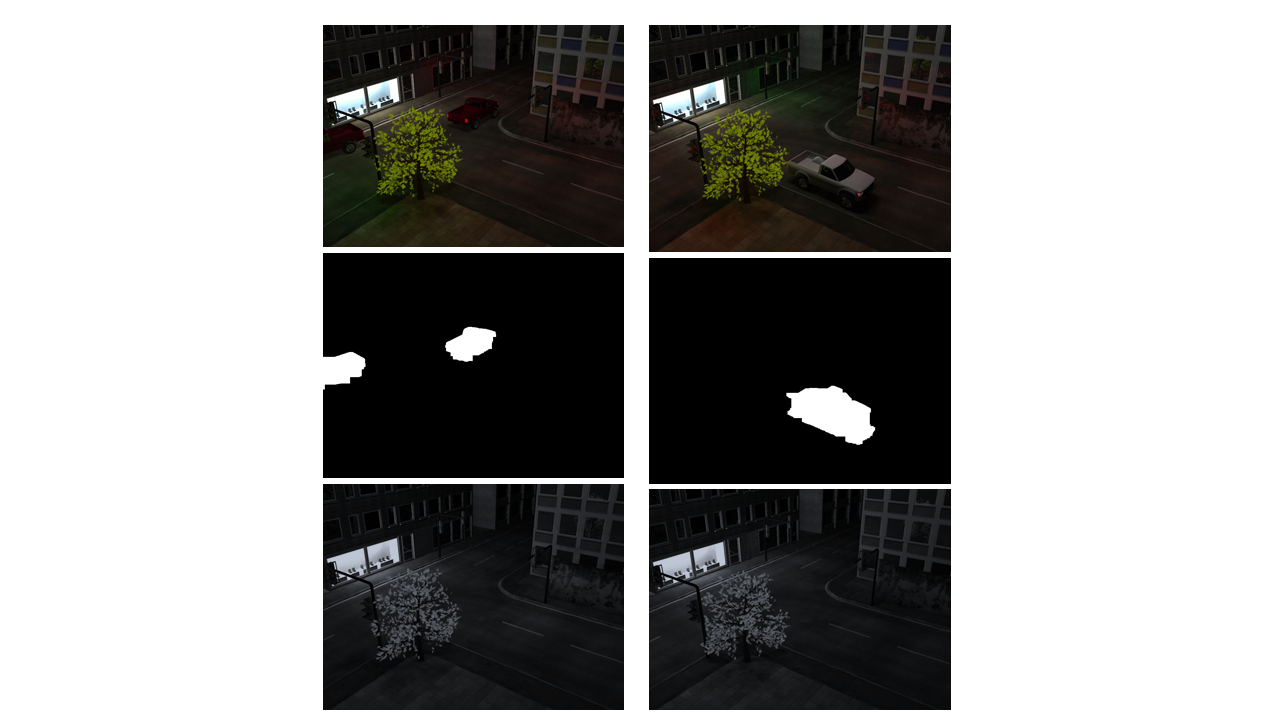}} 
  \caption{Result of applying the three methods (a) RPCA, (b) RMC, and (c) fRMC on SABS light switch video demonstrated for two different frames; first row is the original frame, second row is the binary mask of the foreground by thresholding the detected foreground, and last row is the background image. As you can see RMC fails in modeling the background in abrupt change of light.}
\label{fig:sabs}
\vspace{-.1in}
\end{figure*}
}
\newcommand{\tablresultBMC}{
\begin{table*}[]  
\small
\begin{center}  
\caption{Computational time and benchmark metrics for the background subtraction task of our fRMC algorithm compared to RPCA and RMC evaluated on the BMC dataset.}
 \begin{tabular}{| c | c | c | c | c | c | c |} 
 \hline
   Algorithm & Running time (min) & F-measure & D-score & Precision & PSNR & Ssim \\ 
 \hline
  \rowcolor[gray]{0.8}
 \multicolumn{7}{|c|} {Big trucks}  \\
 \hline
 RPCA        &  18           & 0.68  & 0.010 & 0.80 & 26.98 & 0.92 \\ 
 \hline
 RMC         &  29.6         & 0.76   & 0.0088  & 0.93 & 29.0 & 0.93  \\ 
 \hline
\text{fRMC}  &  \textbf{8.4} & \textbf{0.79} & \textbf{0.0087} & \textbf{0.94} & \textbf{30.07} & \textbf{0.94} \\ 
 \hline
   \rowcolor[gray]{0.8}
 \multicolumn{7}{|c|}{Wandering students} \\
 \hline
 RPCA  &  6.2 & 0.87 & 0.0062 & 0.90 & 46.03 & 0.97 \\
 \hline
 RMC &  12 & 0.80  & 0.0094 & \textbf{0.94}  & 42.93 & 0.97 \\ 
 \hline
 \text{fRMC}  &  \textbf{3.4} & \textbf{0.87} & \textbf{0.0061} & 0.90 & \textbf{46.11} & 0.97 \\
 \hline
   \rowcolor[gray]{0.8}
 \multicolumn{7}{|c|}{Rabbit in the night} \\
 \hline
 RPCA &  28 & 0.60 & 0.0053 & 0.61 & 41.19 & 0.97 \\ 
 \hline
 RMC &  30.3 & \textbf{0.76}  & \textbf{0.0035} & \textbf{0.91}  & \textbf{48.04}  & 0.98 \\ 
 \hline
 \text{fRMC}  &  \textbf{8.6} & 0.75 & 0.0037 & 0.85 & 47.42 & \textbf{0.99} \\
 \hline
   \rowcolor[gray]{0.8}
 \multicolumn{7}{|c|}{Beware of the trains} \\
 \hline
 RPCA  & 11.46 & 0.683 & 0.105 & 0.778 & 33.348 & 0.95 \\
 \hline
 RMC &  16 & 0.64  & 0.0091 & 0.83 & 32.94 & 0.95 \\ 
 \hline
 \text{fRMC} & \textbf{4.3} & \textbf{0.71} & \textbf{0.0084} & \textbf{0.89} & \textbf{34.3} & \textbf{0.96} \\
\hline
  \rowcolor[gray]{0.8}
\multicolumn{7}{|c|}{Train in the tunnel} \\
 \hline
 RPCA  & 14.2 & 0.63 & \textbf{0.0076} & 0.81 & 27.18  & 0.93  \\
 \hline
 RMC &  26.8 & \textbf{0.78}  & 0.0085  & 0.90 & \textbf{30.59} & \textbf{0.94} \\ 
 \hline
 \text{fRMC}  & \textbf{7.8} & \textbf{0.78}  & 0.0086 & \textbf{0.91} & \textbf{30.59} & \textbf{0.94} \\
\hline

\end{tabular}
\label{tbl:BMCresults}
\end{center}
\vspace{-.2in}
\end{table*}
}

\newcommand{\tablresultSABS}{
\begin{table}[]  
\small
\begin{center}  
\caption{Computational time and benchmark metrics for the background subtraction task of our fRMC algorithm compared to RPCA and RMC evaluated on the SABS datasets.$^\star$shows the best result for each experiment reported by VIS for SABS dataset.}
 \begin{tabular}{| c | c | c | c |} 
 \hline
   Algorithm & Running time (min) & Precision & F-measure  \\ 
 \hline
  \rowcolor[gray]{0.8}
 \multicolumn{4}{|c|} {Basic}  \\
 \hline
 RPCA & 9.2  & \textbf{0.734}  &  0.775   \\ 
 \hline
 RMC & 45.0   & 0.732   & 0.778  \\ 
 \hline
\text{fRMC}  &  \textbf{6.2} &  0.732  & 0.779 \\ 
 \hline
 Best result$^\star$ & - & - & \textbf{0.800} \\
 \hline
   \rowcolor[gray]{0.8}
 \multicolumn{4}{|c|}{Light switch} \\
 \hline
 RPCA  & 9.0  & \textbf{0.725} & 0.506  \\
 \hline
 RMC & 44.0  & 0.100  & 0.163  \\ 
 \hline
 \text{fRMC}  & \textbf{6.3}  & 0.705 & \textbf{0.495}  \\
 \hline
 Best result & - & - & 0.316 \\
 \hline
   \rowcolor[gray]{0.8}
 \multicolumn{4}{|c|}{Camouflage} \\
 \hline
 RPCA & 8.8  & 0.742 & 0.747 \\ 
 \hline
 RMC & 45  & 0.746  & 0.752  \\ 
 \hline
 \text{fRMC}  & \textbf{5.6}   & \textbf{0.746} & 0.753  \\
 \hline
 Best result & - & - & \textbf{0.820} \\
 \hline
   \rowcolor[gray]{0.8}
 \multicolumn{4}{|c|}{No camouflage} \\
 \hline
 RPCA  & 9.6 & \textbf{0.752} & 0.7649 \\
 \hline
 RMC & 44.3  &  0.751 & 0.769  \\ 
 \hline
 \text{fRMC} & \textbf{6.0}  & 0.750  & 0.769   \\
\hline
Best result & - & - & \textbf{0.829} \\
 \hline
  \rowcolor[gray]{0.8}
\multicolumn{4}{|c|}{Noisy night} \\
 \hline
 RPCA  & 9.0  & \textbf{0.889} & 0.539  \\
 \hline
 RMC & 44.2  & 0.725  & 0.527  \\ 
 \hline
 \text{fRMC}  & \textbf{5.1}  & 0.726 & \textbf{0.528} \\
\hline
Best result & - & - & 0.321 \\
 \hline
 \rowcolor[gray]{0.8}
\multicolumn{4}{|c|}{MPEG4 (40Kbps)} \\
 \hline
 RPCA  & 9.5 & 0.662 & \textbf{0.779}   \\
 \hline
 RMC & 44.6  & \textbf{0.715} & 0.777  \\ 
 \hline
 \text{fRMC}  & \textbf{5.4}  & \textbf{0.714}   & 0.776 \\
\hline
Best result & - & - & 0.774 \\
 \hline

\end{tabular}
\label{tbl:SABSresults}
\end{center}
\vspace{-.2in}
\end{table}
}

\usepackage[pagebackref=true,breaklinks=true,letterpaper=true,colorlinks,bookmarks=false]{hyperref}

 \iccvfinalcopy 

\def\iccvPaperID{****} 

\ificcvfinal\fi
\begin{document}

\title{Background Subtraction via Fast Robust Matrix Completion}

\author{Behnaz Rezaei and Sarah Ostadabbas\\
Augmented Cognition Lab (ACLab)\\
Electrical and Computer Engineering Department\\
Northeastern University, Boston, MA 02115\\
{\tt\small brezaei@ece.neu.edu, ostadabbas@ece.neu.edu}
}

\maketitle

\begin{abstract}
Background subtraction is the primary task of the majority of video inspection systems. The most important part of the background subtraction which is common among different algorithms is background modeling. In this regard, our paper addresses the problem of background modeling in a computationally efficient way, which is important for current eruption of "big data" processing coming from high resolution multi-channel videos. Our model is based on the assumption that background in natural images lies on a low-dimensional subspace. We formulated and solved this problem in a low-rank matrix completion framework. In modeling the background, we benefited from the in-face extended Frank-Wolfe algorithm for solving a defined convex optimization problem. We evaluated our fast robust matrix completion (fRMC) method on both background models challenge (BMC) and Stuttgart artificial background subtraction (SABS) datasets. The results were compared with the robust principle component analysis (RPCA) and low-rank robust matrix completion (RMC) methods, both solved by inexact augmented Lagrangian multiplier (IALM). The results showed faster computation, at least twice as when IALM solver is used, while having a comparable accuracy even better in some challenges, in subtracting the backgrounds in order to detect moving objects in the scene. 

\end{abstract}

\section{Introduction}
Background subtraction or foreground detection is the principle task of almost every video inspection algorithm before operating further processing designed for a particular computer vision application. In essence, a robust background subtraction is achieved by creating a model of the background, as depicted in \figref{bs}. The performance of the background subtraction algorithm depends on how well the background model can adapt to the slight and sudden changes of the background scenes. Recently, a huge body of work in this area has focused on modeling the background as a low-dimensional subspace in the high dimensional space of video frames. 

Following the modeling of the background as a low-dimensional subspace, there are various ways to formulate the task of background subtraction as an optimization problem, which has been employed in different works \cite{candes2011robust,guan2012mahnmf,he2012incremental,zhou2013moving}. Although these methods perform well in background/foreground separation, they are designed based on a quadratic optimization problem with heavy structural properties in their iterations. Therefore, the issue of low speed and high memory usage of these methods makes them impractical for online video inspection and prevents them from scaling to big data domain.

\figbs

\subsection{Prior Works in Background Modeling}
The most common modeling of the background is based on the probabilistic modeling, which is first proposed in Stauffer and Grimson \cite{stauffer1999adaptive} and then modified by Hayman and Eklundh \cite{hayman2003statistical}. In this method, which is called Gaussian mixture model (GMM), distribution of the pixel colors is estimated by sum of different Gaussian distributions and the parameters of each distribution are learned through an online expectation maximization (EM) algorithm. Although GMM can manage the slight changes in the background illumination, immediate variations of the background often appear in the foreground. Moreover, if the initial video frames used for the parameter learning are noisy, the trained model will substantially suffer from the noise. In order to improve the performance of the GMM approach, there are various works trying to make this algorithm more robust to noise by proposing different learning methods or modifying the adaptation of the algorithm \cite{chen2007efficient,tuzel2005bayesian,zivkovic2006efficient}. Beside GMM, fuzzy and neuro-fuzzy background subtraction methods are also classified as probabilistic background modeling. Authors in \cite{Sobral20144} provided a comprehensive review of these methods and their performance evaluation. 

Another class of algorithms for modeling the background is based on the implicit or explicit decomposition of an observation matrix into a low-rank matrix and an additive part. Robust principle component analysis (RPCA) was the first problem formulation for this matrix decomposition, which decomposes a matrix into a superposition of a low-rank and a sparse matrix \cite{candes2011robust}. Inspired by RPCA, various problem formulations for decomposing a matrix into a low-rank matrix and an additive matrix were proposed such as non-negative matrix factorization (NMF), robust subspace tracking (RST), robust matrix completion (RMC), and robust low-rank minimization (RLRM) \cite{guan2012mahnmf,he2012incremental,zhou2013moving, 6854862}. The main difference between these methods is the optimization problem defined by each method and the solver that they used to tackle the problem. Practically, in all of the above methods, background sequence is modeled as the low-rank/low-dimensional subspace and the moving objects form the outliers, which are represented by an additive matrix to the low-rank matrix. Although these algorithms work visually well in modeling the background and its gradual changes over time, they are based on an optimization problem that requires to be solved by computationally expensive iterations. Despite various efforts on reducing their computational complexity, it seems that none of them are able to address the challenges of large scale datasets in videos from real-world applications. A survey on background subtraction algorithms with low-dimensional subspace learning framework can be found in \cite{bouwmans2017decomposition, bouwmans2016handbook, Sobral2015}. below, a short review on the state-of-the-art works regarding the above mentioned methods is provided.

Initial work on RPCA as one of the methods based on decomposing a matrix into a low rank and a sparse matrix was developed in parallel by three research groups \cite{candes2011robust, wright2009robust, chandrasekaran2011rank}. They all addressed this problem under minimal assumption by solving a convex optimization problem called principle component pursuit (PCP). After the initial effort, many algorithms have been proposed to solve the PCP making an attempt to reduce the computation and memory cost of the problem \cite{lin2011linearized, lin2009fast, yuan2009sparse, shen2014augmented}. Inexact augmented Lagrangian multiplier (IALM) is a successful effort in this regard for solving the problem of RPCA based on the augmented Lagrangian multiplier (ALM) algorithm without taking the unnecessary singular value decomposition (SVD) steps \cite{lin2010augmented}.

Works categorized in NMF framework are constructed on decomposing a matrix into a product of two non-negative matrices. Conventional methods of NMF are designed to model the Gaussian or Poisson distributions, so they try to fit one of two mentioned distribution to the background and foreground. Consequently, they do not perform well for background/foreground separation, where their distribution has thick tail. Nonetheless, there have been works improving the performance of NMF algorithms for background modeling using different distance/loss functions \cite{guan2012mahnmf, kumar2015near, kumar2013fast, woo2014robust}. RST algorithms try to address tracking of non-stationary subspaces or in other words, online separation of the foreground and background. Associated subspaces can have low-rank or sparse structures similar to previously discussed methods. Following the idea of RST, there are different algorithms, which try to separate background and foreground iteratively using incremental gradient descent constrained on Grassmannian manifolds \cite{he2012incremental, xu2013gosus}. Likewise, both RMC and RLRM methods work based on representing the background as a low-rank matrix.

\subsection{Our Contribution}
Natural videos with static background as the main region of the frames due to the high correlation among frames can be modeled as a low-rank matrix with gross perturbations. Based on this assumption, we propose a method, named fast robust matrix completion (fRMC) to model the background in the framework of matrix completion in order to detect the foreground without any prior knowledge about the moving objects. To recover the low-dimensional subspace spanning the background, we formed a quadratic optimization problem, as stated in \eqnref{7}. For solving the corresponding convex optimization problem, which results in a low-dimensional representation of the background, we benefited from the computationally efficient algorithm of in-face extended Frank-Wolfe method proposed in \cite{freund2017extended}. Changing the problem formulation and using this solver resulted in more than two times \emph{faster} computation, while preserving the performance compared to the RPCA with IALM solver. 

In our proposed fRMC, designed based on the in-face extended Frank-Wolfe solver, iterations start from a rank-1 matrix and SVD gets updated in each iteration based on the previous SVD. In this procedure rank of the updated matrix increases in each iteration by at most one. In contrast, IALM which is currently known as the efficient solver for the RPCA and matrix completion (MC) problems computes a partial SVD in each iteration. Partial SVD returns the singular values of a high rank matrix greater than a specific value. The rank of the updated matrix in IALM monotonically increases with higher slope in each iteration. Since we do not need to calculate SVD in each iteration, the fRMC is considerably faster in high-dimensional data processing. Furthermore, there is no need to store the whole matrix representing the background. Instead, we only have to store the left and right hand side singular vectors with singular values of a low-rank matrix. 

Underlying idea inherited from the original Frank-Wolfe method for solving a convex quadratic optimization is solving a linear optimization sub-problem in each iteration to update the next iteration. In-face extended Frank-Wolf makes the iterations as low as possible, while keeping the error in objective function in desired level by proposing the in-face direction in updating phase. Updating the next iteration toward in-face direction without increasing the rank makes this solver efficient in scaling to huge-size convex optimization, which has shown increasing interest in the computer vision field. 

We evaluated the computational complexity and functionality of our proposed fRMC compared to RPCA and RMC both solved with IALM. Evaluation were according to two publicly available datasets including the background models challenge (BMC 2012) and the Stuttgart artificial background subtraction (SABS) datasets.

\section{Methodology}

\emph{Problem Formulation:} Given an observation matrix $\mathbf{V}$ consisting of each video frame as its columns, the objective is to recover the underlying low-rank matrix, $\mathbf{B}$, from the foreground perturbations $\mathbf{F}$ as: 
\begin{align}
\label{eqn:formulation}
    \min ~~~~ rank(\mathbf{B}) ~~~~\text{s.t.} ~~~~\mathbf{V} = \mathbf{B} + \mathbf{F}
\end{align}
where the recovered low-rank matrix $\mathbf{B}$ works as the background model for the observed video. In the following, we first describe the way that we implicitly separate the moving objects from the background using the framework of the matrix completion as presented in \secref{MC}. This new formulation allows us to employ a more efficient solver with lower iterations as well as lower computations in each iterations, which results in our computationally efficient fRMC method. The details of this solver called in-face extended Frank-Wolf is given in \secref{Frank-Wolf}.
In \secref{RPCA}, to provide a comparison platform, we then give an overview on the RPCA and how it is used for modeling the background using explicit decomposition of a matrix into a low-rank and sparse matrix. RPCA is the most common formulation for the low-rank representation of the background when it is solved with PCP \cite{candes2011robust}. As RPCA has shown promising results for the background subtraction, we implemented this method using the IALM solver introduced in \cite{lin2010augmented} and compared results of our proposed problem formulation with this algorithm. In addition, in order to compare the performance and speed of our suggested solver with IALM for the same optimization problem, we 
used IALM for our problem formulation in \eqnref{formulation} in a form of a conventional robust matrix completion (RMC) algorithm. The performance evaluation and comparison results are given in \secref{results}.

\subsection{Matrix Completion Framework} 
\label{sec:MC}
One way to think about the background subtraction problem is to consider the background throughout the video as a collection of high-dimensional data lying in a low-dimensional subspace due to the high correlation of the backgrounds among video frames. So far, most of the optimization problems formulated for modeling the background as a low-rank matrix have imposed the sparse structure to the representing matrix of the foreground by $l_1$-norm minimization. However, it is not the only way to impose sparsity. As it is theoretically and experimentally shown, under mild conditions, Frobenius norm can be effective in imposing sparsity  \cite{peng2016connections, peng2016constructing, peng2016automatic}. Considering this theory, we modeled the background as a low-dimensional subspace and implicitly impose sparsity to the foreground by minimizing the Frobenius norm, which is strictly convex compared to using $l_1$-norm minimization.
In order to form our optimization problem we used the low-rank matrix completion (LRMC) concept.

In LRMC, we have a partially observed data matrix $\mathbf{D} \in \mathbb{R}^{n_1,n_2} $ containing $n_2$ observations, each of dimension $n_1$, which only a fraction of its entries are available and the task is to predict unobserved entries of the $\mathbf{D}$ matrix. Matrix $\mathbf{Z}\in \mathbb{R}^{n_1,n_2}$ is the estimated matrix of $\mathbf{D}$ containing the prediction of unobserved entries. In general, without any assumption on the underlying structure of the estimated matrix $\mathbf{Z}$, this problem is ill-posed and the data matrix can be filled up with any real values. The most common assumption without having any prior knowledge about the data distribution is to restrict matrix $\mathbf{Z}$ to a low-rank matrix \cite{freund2017extended}. This assumption is accurate in many real applications with high dimensionality, including background subtraction or moving object detection. The relaxed optimization problem of the LRMC is stated as:
\begin{align} \label{eqn:6}
    \min~f(z):=1/2\sum_{(i, j) \in \Omega}(z_{ij}-d_{ij})^2 ~~~~ \text{s.t.} ~~~ ||\mathbf{Z}||_{\star} < \delta
\end{align}
where  $f(\cdot)$ is the least squares error between the available observed data entries $d_{ij}$ and estimated data entries $z_{ij}$, and $\Omega$ is the subset containing all indices of observed entries. $||~.~||_{\star}$ indicates the nuclear norm of a matrix, which is defined as sum of its singular values, and $\delta$ is the constraining upper band for the nuclear norm ball of estimated low-rank matrix $\mathbf{Z}$. It is assumed that the observed entries are contaminated by noise with bounded energy and minimizing the $||\mathbf{Z}||_{\star}$ imposes the low-rank constraint on the estimated matrix.

Although in the background subtraction context we do not have any missing data, instead we want to recover the background from observations corrupted by foreground (as noise). Considering the fact that background is lying on a low-dimensional subspace compared to the space domain of the video frames, task of the background subtraction can be restated in the LRMC framework. If data matrix $\mathbf{V}$ is formed by arranging the video frames in each column of it, in order to recover the background from video frames, we estimate the low-rank matrix $\mathbf{B}$ out of the uncontaminated entries of the observed video $\mathbf{V}$. On the other hand, foreground is the difference of the estimated background video and the original video, which in the LRMC is the sparse bounded energy perturbation of the background. Low-rank background model in the context of the LRMC is restated in solving the following optimization problem:
\begin{align} \label{eqn:7}
     \min ~~ ||\mathbf{B}-\mathbf{V}||^2_F ~~~~ \text{s.t.} ~~~ ||\mathbf{B}||_{\star}< \delta   
\end{align}
where $||~.~||_F$ indicates the Frobenius norm of a matrix defined as root sum squared of matrix entries, and $(\mathbf{B}-\mathbf{V})$ is related to the foreground perturbations in videos.

\subsection{Extended Frank-Wolfe Method for LRMC}
\label{sec:Frank-Wolf}
Basically, LRMC optimization formulation is a specific case of RPCA objective function, therefore this problem can be solved using the IALM method. However, IALM has scaling problem to massive datasets, which arise in image/video processing. Multi-channel videos usually have millions of dimensions, hence implementation of IALM, considering that it requires taking an SVD in each iteration, is not a feasible solution. One of the algorithms designed for convex problems with better scalability is Frank-Wolfe method and its extensions featuring linear convergence \cite{guelat1986some,freund2017extended, mu2016scalable}. In order to recover the substantial low-rank matrix $\mathbf{B}$, which represents the background throughout video frames, we have used in-face extended Frank-Wolfe algorithm proposed in \cite{freund2017extended}. 

In fact, \eqnref{6} is an example of a more general problem as below, which is addressed by Frank-Wolf method:
\begin{align} \label{eqn:8}
    f^{\star}:= \text{min}~~~\underset{x \in S}{f(x)}
\end{align}
where $S$ is a closed and bounded convex set, and $f(\cdot)$ is a differentiable convex function defined on subset $S$. Original Frank-Wolfe method finds the sub-optimal solution for this problem based on the procedure described in \algref{2}.

\begin{algorithm}[t]
\SetAlgoLined
\KwResult{$x^{\star}$ }
 initialization: $x_0,~ \text{lower bound:}~ C_{-1},~ i \leftarrow 0$ \;
 \While{not converged}{
  1. compute $\nabla f(x)$\;
2. $\tilde{x}_i \leftarrow argmin\{f(x_i) + \nabla f{(x_i)}^T (x-x_i)\}$ \;
3. $C^{\omega}_i \leftarrow f(x_i) + \nabla f{(x_i)}^T (\tilde{x}-x_i)$\;
//~updating the best bound\\
4. $C_i \leftarrow \max\{C_{i-1},C_i^{\omega}\}$\;
//~updating $x$\\
 5. $x_{i+1} \leftarrow x_i + \bar{\alpha}_i(\tilde{x_i}-x_i), ~ \bar{\alpha}_i \in [ 0, 1] $\;
 }
\caption{Original Frank-Wolfe method for the optimization problem in \eqnref{8}.}  \label{alg:2}
\end{algorithm}
$C_i$ is the optional lower bound for the optimal objective function $f^{\star}$ and is updated in steps 3 and 4 of each iteration. This lower bound is effective in conditioning the number of iterations. $\bar{\alpha_i}$ is the updating step size, which can be selected by line-search or simple rule of $\bar{\alpha_i}:=\frac{2}{i+1}$. There are strong mathematical verification for the computational guarantee of Frank-Wolf method \cite{freund2016new}. The main computational load of the Frank-Wolf algorithm is solving the linear optimization problem in step $2$ of \algref{2}. 

\begin{algorithm}[t]
\caption{In-face extended Frank-Wolfe method for the optimization problem in \eqnref{7}.}  \label{alg:3}
\SetAlgoLined
\KwResult{$\mathbf{B}^{\star}$ }
 Input: $\mathbf{V}$\;
 Constants: $ 0 \leq \gamma_1 \leq \gamma_2 \leq 1 , ~ \bar{L} \geq L=1,~ \bar{D} \geq D = 2 \delta$ \;
 Definition: 
 $f(\mathbf{B}^i) = 1/2 ||\mathbf{B}^i-\mathbf{V} ||^2_{F}$ ,\\
 ~~~~~~~~~~~~~~~~$\nabla f(\mathbf{B}^i)=(\mathbf{B}^i-\mathbf{V}),~ \mathbf{B} \bullet \mathbf{X} := tr\{\mathbf{B}^T \mathbf{X} \}$ \;
 Initialization: $\mathbf{B}^0 \leftarrow -\delta u_0 v^T_0, i \leftarrow 0$\\
 $~~~~~~~~~~~~~~~~~~~ \text{lower bound}:C_{-1} \leftarrow \max \{f(0)+\nabla f(0) \}$\;
 \While{not converged}{
1. $\nabla f(\mathbf{B}^i) = (\mathbf{B}^i -\mathbf{V}),~~ C_i \leftarrow C_{i-1}$ \;
2. compute direction $q^i$ \\
~~~~ $\hat{\mathbf{B}}^i \leftarrow argmax~ \nabla f(\mathbf{B}^i) \bullet \mathbf{B} ;~\mathbf{B} \in \mathfrak{F}_{C}(\mathbf{B}^i)$ \\
~~~~ $q^i \leftarrow \mathbf{B}^i - \hat{\mathbf{B}}^i$ \\

~~~~\textbf{case 1}: in the case 
$\mathbf{B}^i \in int({C})$ and when $\mathbf{B}^i \in \partial ({C}) \Rightarrow \mathfrak{F}_{{C}}(\mathbf{B}^i)={C}$\\
$~~~~~~~~~~\hat{\mathbf{B}}^i = \delta \mathbf{u}_i \mathbf{v}^T_i$  \;
~~~~\textbf{case 2}: in the case
$\mathbf{B}^i \in \partial({C}),~ rank(\mathbf{B}^i) =r \Rightarrow \mathbf{B}^i=\mathbf{UDV}^T$ \\
$~~~~~~~~~~ \hat{\mathbf{B}}^i \leftarrow \mathbf{U}\hat{\mathbf{M}}^i\mathbf{V}^T = \delta \mathbf{Uu}_i \mathbf{u}^T_i \mathbf{V}^T$ \\
3. compute step size\\
$~~~~~\alpha ^{stop}_i \leftarrow \{ \alpha : \mathbf{B}^i +\alpha q^i \in \mathfrak{F}_{C}(\mathbf{B}^i) \}$\;
~~~~~$\mathbf{B}^i_b := \mathbf{B}^i + \alpha ^{stop}_i q^i $\;
~~~~~$\mathbf{B}^i_a := \mathbf{B}^i + \bar{\beta}_i q^i ~\text{where}~ \bar{\beta}_i \in [0, \alpha ^{stop}_i]$\;
4. choose next iterate:\\
~~~~~(a) if $ \frac{1}{f(\mathbf{B}^i_{b})-C_i} \geq \frac{1}{f(\mathbf{B}^i)-C_i}+\frac{\gamma_{1}}{2\bar{L}\bar{D}^2}$:\\
~~~~~~~~~~~~$\mathbf{B}^{i+1} \leftarrow \mathbf{B}^i_a$\;
~~~~~(b) if $\frac{1}{f(\mathbf{B}^i_{a})-C_i} \geq \frac{1}{f(\mathbf{B}^i)-C_i}+\frac{\gamma_{2}}{2\bar{L}\bar{D}^2}$: \\
~~~~~~~~~~~~$\mathbf{B}^{i+1} \leftarrow \mathbf{B}^i_b$\;
~~~~~(c) Else:\\
~~~~~~~~~~~~ do the original Frank-Wolfe step with the \\
~~~~~~~~~~~~ input $x_i = \mathbf{B}^i$ and update lower bound \\ 
~~~~~~~~~~~~ described in \algref{2}\;
 }
\end{algorithm}

In this work, we used in-face extended Frank-Wolfe method to solve the problem of foreground detection introduced in \eqnref{7}. This method computes and works with points that have specific structure of low-rank in the case when $x$ is a data matrix. The main advantage of this algorithm toward optimizing the objective function $f(x)$ is iterating in low-dimensional faces of $S$. Such low-rank iteration not only make the output solution low-rank, but also result in an essential reduction in computation cost. This method adopted for the specific problem of foreground detection in \eqnref{7} is presented in \algref{3}. In this algorithm, $\mathbf{B}^i = \mathbf{UDV}^T$ denotes the SVD of the current iteration, and $\mathbf{u}_i$ and $\mathbf{v}_i$ are the left and right singular vectors corresponding to the largest singular values of the matrix $\nabla{f(\mathbf{B}^i)}$. At step 2 of each iteration, algorithm works for the "in-face" direction $q^i$ which preserves the next estimated point in the minimal face $\mathfrak{F}_C(\mathbf{B}^i)$. In step 4, algorithm chooses between three possible next steps: $\mathbf{B}^i_b$ that lies in the relative boundary of the current minimal face, $\mathbf{B}^i_a$ that may not lie in the relative boundary of the current minimal face, and a regular Frank-Wolf update in step 4(c). Each of these steps are chosen based on the decrease in the optimality bound gap criterion:
\begin{align} \label{eqn:9}
    \frac{1}{f(\mathbf{B}^i_{b/a})-C_i} \geq \frac{1}{f(\mathbf{B}^i)-C_i}+\frac{\gamma_{1/2}}{2\bar{L}\bar{D}^2}
\end{align}
where $\mathbf{B}^i_a$ and $\mathbf{B}^i_b$ are the two in-face candidates for the next update of the $\mathbf{B}^i$. $\bar{L}$ is the Lipschitz constant of the gradient of $f(.)$ on nuclear norm ball, and $\bar{D}$ is the diameter of the convex bounded set in the optimization problem, which in our case is the diameter of the nuclear norm ball equal to $2\delta$. $\gamma_1$ and $\gamma_2$ are two constants that control the convergence properties of the algorithm. When we want the criterion in \eqnref{9} be easily satisfied particularly in low-rank matrix completion, $\gamma_1$ and $\gamma_2$ are preferred to be lower than higher. 

Defining the problem of background subtraction as the convex optimization problem in \eqnref{7} and benefiting from high performance in-face extended Frank-Wolf with verified convergence properties to solve the proposed problem configure our fRMC method.

\subsection{An Overview of Robust Principle Component Analysis}
\label{sec:RPCA}
Under two assumptions that the low-rank matrix is not sparse and sparsity pattern of the sparse component is selected randomly, RPCA can decompose a matrix into a superposition of a low-rank matrix and a sparse matrix. Despite the differences between RPCA and matrix completion, the recovery of the low-rank matrix in the RPCA can be treated as a low-rank matrix completion. In this case, the matrix completion deviates from recovering a low-rank matrix from a fraction of its components to recovering a low-rank matrix by having an unknown fraction of them available, while the rest of them are grossly corrupted \cite{candes2011robust}. J. Candes in \cite{candes2011robust} approaches the problem of recovering the incomplete and corrupted entries as a convex optimization:
\begin{align} \label{eqn:1}
 \min ~~~||\mathbf{B}||_{\star}+\lambda ||\mathbf{F}||_1~~~~\text{s.t.} ~~~ \mathbf{V}=\mathbf{B}+\mathbf{F} 
\end{align}
where $||~.~||_1$ is $l_1$-norm of a matrix defined as the sum of absolute values of matrix entries, and $\lambda$ is a positive regularizing coefficient. 

Efficient algorithm for this optimization is augmented Lagrangian multiplier (ALM) introduced in \cite{yuan2009sparse,lin2011linearized}. ALM algorithm inspired from the theory that first-order iterative thresholding algorithms can be used efficiently for both $l_1$-norm and nuclear-norm minimization \cite{cai2010singular, beck2009fast, cai2009linearized}. 
The original ALM is designed for solving the constrained optimization problem of:
\begin{align} \label{eqn:2}
\min ~ f(X)~~~~~~ \text{s.t.}~~~~~~ g(X)=0 
\end{align}
where $f:\mathbb{R}^{n_1}\rightarrow \mathbb{R}$ and $g:\mathbb{R}^{n_1} \rightarrow \mathbb{R}^{n_2}$. ALM algorithm converts constrained problem in \eqnref{2} to an unconstrained problem by adding a penalty term that punishes violations from the equality constraint:
\begin{align} \label{eqn:3}
L_{\rho}(X, Y) = f(X) +\langle \mathbf{Y}, g(X) \rangle +\frac{\rho}{2} ||g(X)||^{2}_{F} 
\end{align}
where $\mathbf{Y}$ is a regularizer matrix, $\langle \mathbf{Y}, g(X) \rangle$ is defined as $trace(\mathbf{Y}^T\times g(X))$ and $\rho$ is a positive scalar. ALM algorithm formulation for the RPCA problem is identified as:
\begin{align} \label{eqn:4}
\begin{split}
X=(\mathbf{B},\mathbf{F}), ~~~~~  f(X) &= ||\mathbf{B}||_{\star}+\lambda ||\mathbf{F}||_1,\\
g(X) &= \mathbf{V}-\mathbf{B}-\mathbf{F}.    
\end{split}
\vspace{-0.05in}
\end{align}
then:
\vspace{-0.05in}
\begin{align} \label{eqn:5}
\begin{split}
&L_{\rho}(\mathbf{B}, \mathbf{F}, \mathbf{Y}) = \\ &||\mathbf{B}||_{\star}+\lambda ||\mathbf{F}||_1 +\langle \mathbf{Y}, \mathbf{V}-\mathbf{B}-\mathbf{F} \rangle +\frac{\rho}{2} ||\mathbf{V}-\mathbf{B}-\mathbf{F}||^{2}_{F}
\end{split}
\end{align}
The state-of-the art solver for the unconstrained optimization problem in \eqnref{5} in terms of speed and computational guarantee is the IALM, which its major computational cost is a partial SVD in each iteration. 
Considering $n$ as the $\max{(n_1,n_2)}$, it is shown that for incoherent $\mathbf{B}$, correct recovery occurs with high probability when the rank of $\mathbf{B}$ is in the order of ${n}/{\mu {(\log{n})^2}}$, where $\mu >1$  and the number of nonzero elements in $\mathbf{F}$ is on the order of $n^2$. A choice of $\lambda = 1/\sqrt{n}$ works with high probability for recovering the low-rank matrix \cite{lin2011linearized, lin2009fast}.

As we pointed out earlier, IALM solver can be used for solving the unconstrained optimization problem related to the low-rank matrix completion as a special case of the objective function of RPCA demonstrated as:
\begin{align} \label{eqn:14}
\begin{split}
L_{\rho}(\mathbf{B}, Y) = ||\mathbf{B}||_{\star}+\langle \mathbf{Y}, \mathbf{V}-\mathbf{B} \rangle +\frac{\rho}{2} ||\mathbf{V}-\mathbf{B}||^{2}_{F}
\end{split}
\end{align}

\section{Experimental Analysis}
\label{sec:results}

\subsection{Performance Evaluation on BMC Dataset} 
In order to validate and compare our proposed fRMC formulation in terms of computational time and detection performance, we tested fRMC, RPCA, and RMC on the BMC 2012 dataset \cite{vacavant2012benchmark}. This dataset contains 9 real videos included with encrypted ground-truth images of the foreground to test the background models. The software given by the website (called BMCWizard \footnote{\url{http://bmc.iut-auvergne.com/?page_id=63}}) measures two types of metrics for performance evaluation: static quality metrics and application quality metrics. 
Regarding the static quality two scores are measured in this application: F-measure and peak signal-noise ratio (PSNR). These two benchmarks are introduced to evaluate the raw behavior of each algorithm for moving object segmentation. Considering $\mathbf{F}$ as the set of $n$ foreground frames processed by the background subtraction algorithm, and $\mathbf{G}$ as the ground-truth images, these two metrics are defined as follows:
\begin{align} \label{eqn:10}
\begin{split}
    & F = \frac{1}{n}  \sum_{i=1}^{n}2 \frac{Prec_i \times Rec_i}{Prec_i + Rec_i}
    ~~~~~~~~~~~~\text{where}:\\
    & Rec_i(P) = \frac{TP_i}{TP_i + FN_i} ;
    ~~~Prec_i(P) = \frac{TP_i}{TP_i + FP_i};\\
    & Rec_i(N) = \frac{TN_i}{TN_i + FP_i} ; 
    ~~~ Prec_i(N) = \frac{TN_i}{TN_i + FN_i} \\
    & Rec_i = 1/2(Rec_i(P)+ Rec_i(P)); \\
    &Prec_i = 1/2(Prec_i(P) + Prec_i(P)) 
\end{split}
\end{align}
where for given frame $i$, $TP_i$ and $FP_i$ are the true and false positive detection, and $TN_i$ and $FN_i$ are the true and false negative detection. And, 
\begin{align} \label{eqn:11}
    PSNR = \frac{1}{n} \sum_{i=1}^{n} 10 \log_{10} \frac{n_1}{\sum_{j=1}^{n_1} ||\mathbf{F}_i(j) - \mathbf{G}_i(j) ||^2}
\end{align}
where $\mathbf{F}_i(j)$ is the $j$th pixel of the image $i$ (of size $n_1$) in the frame sequence $\mathbf{F}$. For the application quality metrics that consider the problem of background subtraction in a visual and perceptual way, BMCWizard calculates two other benchmark: Structural similarity (SSIM) and D-score. SSIM is calculated as:
\begin{align} \label{eqn:12}
     SSIM(\mathbf{F}, \mathbf{G}) = \frac{1}{n} \sum_{i=1}^{n} \frac{(2 \mu_{\mathbf{F}_i} \mu_{\mathbf{G}_i}+c_1)(2cov_{\mathbf{F}_i \mathbf{G}_i}+c_2)}{(\mu^2_{\mathbf{F}_i}+\mu^2_{\mathbf{G}_i}+c_1)(\sigma ^2_{\mathbf{F}_i}+\sigma^2_{\mathbf{G}_i}+c_2)}
\end{align}
where $\mu_{\mathbf{F}_i}$, $\mu_{\mathbf{G}_i}$ are the means and $\sigma_{\mathbf{F}_i}$, $\sigma_{G_i}$ are the standard deviations and $cov_{\mathbf{F}_i\mathbf{G}_i}$ is the covariance of the $\mathbf{F}_i$ and $\mathbf{G}_i$. $c_1$ and $c_2$ are two constants chosen as $6.5025$ and $58.5225$, respectively. Associated to the application quality, D-score considers localization of errors based on the real object position. To compute this benchmark, only mistakes in the background subtraction algorithm is taken into account:
\begin{align} \label{eqn:13}
    D-score(\mathbf{F}_i(j)) = exp((-\log_2 (2 DT(\mathbf{F}_i(j))-5/2)^2)
\end{align}
where $DT(\mathbf{F}_i(j))$ is given by minimal distance between the pixel $\mathbf{F}_i(j)$ and the nearest reference point (here by Baddeley distance). With this metric, local/far errors will produce a near zero D-score. In contrast, medium range errors produce high D-score; a good D-score tends to zero.

\tablresultBMC

We tested our algorithm on six videos of the BMC dataset in comparison with both RPCA and RMC. The performance metrics calculated by BMCWizard are reported in \tblref{BMCresults}. As demonstrated by the results, with almost the same performance (even better in some metrics) our fRMC algorithm processes the same video in less than half of the time required by RPCA. Experiments were executed in MATLAB installed on a machine with \textit{Intel(R) Xeon(R)@ 3.60GHz, 6 Cores} CPU and $32GB$ RAM. Some examples of background and foreground masks resulted from RPCA, RMC and our fRMC background subtraction methods are shown in \figref{bmc}. As you can see, all outcomes are very comparable in mask detection.

\figbmc

\figsabs

\subsection{Performance Evaluation on SABS Dataset}
\tablresultSABS
As the second part of our performance evaluation we tested the functionality of fRMC, RPCA, and RMC algorithms in background/ foreground separation on the SABS dataset \cite{cvpr11brutzer}. This dataset contains 9 artificial videos covering the typical challenges in the background subtraction. The considered challenges are: 

\textbf{Gradual illumination changes}: to measure the robustness of the background models in gradual changes of the environment (e.g. variation of light intensity in outdoor settings).

\textbf{Sudden illumination changes}: considering the strong changes in the background appearance,  which can cause false positive detection (e.g. sudden switch off).

\textbf{Dynamic background}: taking into account situations with some moving components that are relevant to the background (e.g. traffic lights or trees).

\textbf{Camouflage}:  one of the important challenges in background modeling especially in surveillance applications is similar appearance of some objects to the background, making the precise classification difficult.

\textbf{Shadows}: shadows are irrelevant areas that prevent the classifier to separate the nearby foreground objects as they overlap. 

\textbf{Video noise}: inevitable issue in recording the video is the sensor noise or compression artifact that degrades the signal quality. 

Each video reflects one or more of the above challenges. We tuned the parameters of the algorithms based on the "Basic" video in the dataset and use the same parameters for all the videos.  The benchmark metrics and execution time of each method are reported in \tblref{SABSresults}. The metrics consist of F-measure and precision defined in \eqnref{10} and were measured using the evaluation framework provided by the institute for visualization and interactive systems (VIS)
\interfootnotelinepenalty=10000\footnote{\url{http://www.vis.uni-stuttgart.de/en/research/information-visualisation-and-visual-analytics/visual-analytics-of-video-data/sabs.html}}.
We demonstrated the result of applying the RPCA, RMC, and fRMC on the SABS light switch video in \figref{sabs}. This video covers the challenge of sudden illumination changes and as you can see from the images, RMC cannot model the background satisfactorily and the window is shown as part of the foreground which confirms the low score of RMC in \tblref{SABSresults} for this video.

\section{Conclusion}
In this paper, we proposed our fast robust matrix completion (fRMC)  method to address the problem of time efficient background subtraction by finding the low-dimensional subspace which background lies on that. For recovering the associated subspace, we represented our problem in a matrix completion framework, and benefited from the in-face extended Frank-Wolfe algorithm to solve corresponding optimization problem. We validated our algorithm on the BMC 2012 and SABS datasets in comparison with RPCA and RMC algorithms both solved by IALM solver. With almost the same performance or better in some videos, we accomplished the task with less than half of the execution time required for the RPCA and RMC. 

\vspace{-0.05in}
\balance
\small
\bibliographystyle{ieee}
\bibliography{egbib}

\begin{thebibliography}{10}\itemsep=-1pt

\bibitem{beck2009fast}
A.~Beck and M.~Teboulle.
\newblock A fast iterative shrinkage-thresholding algorithm for linear inverse
  problems.
\newblock {\em SIAM journal on imaging sciences}, 2(1):183--202, 2009.

\bibitem{bouwmans2016handbook}
T.~Bouwmans, N.~S. Aybat, and E.-h. Zahzah.
\newblock {\em Handbook of Robust Low-Rank and Sparse Matrix Decomposition:
  Applications in Image and Video Processing}.
\newblock CRC Press, 2016.

\bibitem{bouwmans2017decomposition}
T.~Bouwmans, A.~Sobral, S.~Javed, S.~K. Jung, and E.-H. Zahzah.
\newblock Decomposition into low-rank plus additive matrices for
  background/foreground separation: A review for a comparative evaluation with
  a large-scale dataset.
\newblock {\em Computer Science Review}, 23:1--71, 2017.

\bibitem{cvpr11brutzer}
{Brutzer, S. and H{\"o}ferlin, Benjamin and Heidemann, Gunther}.
\newblock {Evaluation of Background Subtraction Techniques for Video
  Surveillance}.
\newblock In {\em {Computer Vision and Pattern Recognition (CVPR)}}, pages
  {1937--1944}. {IEEE}, {2011}.

\bibitem{cai2010singular}
J.-F. Cai, E.~J. Cand{\`e}s, and Z.~Shen.
\newblock A singular value thresholding algorithm for matrix completion.
\newblock {\em SIAM Journal on Optimization}, 20(4):1956--1982, 2010.

\bibitem{cai2009linearized}
J.-F. Cai, S.~Osher, and Z.~Shen.
\newblock Linearized bregman iterations for compressed sensing.
\newblock {\em Mathematics of Computation}, 78(267):1515--1536, 2009.

\bibitem{candes2011robust}
E.~J. Cand{\`e}s, X.~Li, Y.~Ma, and J.~Wright.
\newblock Robust principal component analysis?
\newblock {\em Journal of the ACM (JACM)}, 58(3):11, 2011.

\bibitem{chandrasekaran2011rank}
V.~Chandrasekaran, S.~Sanghavi, P.~A. Parrilo, and A.~S. Willsky.
\newblock Rank-sparsity incoherence for matrix decomposition.
\newblock {\em SIAM Journal on Optimization}, 21(2):572--596, 2011.

\bibitem{chen2007efficient}
Y.-T. Chen, C.-S. Chen, C.-R. Huang, and Y.-P. Hung.
\newblock Efficient hierarchical method for background subtraction.
\newblock {\em Pattern Recognition}, 40(10):2706--2715, 2007.

\bibitem{freund2016new}
R.~M. Freund and P.~Grigas.
\newblock New analysis and results for the frank--wolfe method.
\newblock {\em Mathematical Programming}, 155(1-2):199--230, 2016.

\bibitem{freund2017extended}
R.~M. Freund, P.~Grigas, and R.~Mazumder.
\newblock An extended frank--wolfe method with “in-face” directions, and
  its application to low-rank matrix completion.
\newblock {\em SIAM Journal on Optimization}, 27(1):319--346, 2017.

\bibitem{guan2012mahnmf}
N.~Guan, D.~Tao, Z.~Luo, and J.~Shawe-Taylor.
\newblock Mahnmf: Manhattan non-negative matrix factorization.
\newblock {\em arXiv preprint arXiv:1207.3438}, 2012.

\bibitem{guelat1986some}
J.~Gu{\'e}lat and P.~Marcotte.
\newblock Some comments on wolfe's ‘away step’.
\newblock {\em Mathematical Programming}, 35(1):110--119, 1986.

\bibitem{hayman2003statistical}
E.~Hayman and J.-O. Eklundh.
\newblock Statistical background subtraction for a mobile observer.
\newblock In {\em ICCV}, volume~1, 2003.

\bibitem{he2012incremental}
J.~He, L.~Balzano, and A.~Szlam.
\newblock Incremental gradient on the grassmannian for online foreground and
  background separation in subsampled video.
\newblock In {\em Computer Vision and Pattern Recognition (CVPR), 2012 IEEE
  Conference on}, pages 1568--1575. IEEE, 2012.

\bibitem{kumar2015near}
A.~Kumar and V.~Sindhwani.
\newblock Near-separable non-negative matrix factorization with ℓ1 and
  bregman loss functions.
\newblock In {\em Proceedings of the 2015 SIAM International Conference on Data
  Mining}, pages 343--351. SIAM, 2015.

\bibitem{kumar2013fast}
A.~Kumar, V.~Sindhwani, and P.~Kambadur.
\newblock Fast conical hull algorithms for near-separable non-negative matrix
  factorization.
\newblock In {\em ICML (1)}, pages 231--239, 2013.

\bibitem{lin2010augmented}
Z.~Lin, M.~Chen, and Y.~Ma.
\newblock The augmented lagrange multiplier method for exact recovery of
  corrupted low-rank matrices.
\newblock {\em arXiv preprint arXiv:1009.5055}, 2010.

\bibitem{lin2009fast}
Z.~Lin, A.~Ganesh, J.~Wright, L.~Wu, M.~Chen, and Y.~Ma.
\newblock Fast convex optimization algorithms for exact recovery of a corrupted
  low-rank matrix.
\newblock {\em Computational Advances in Multi-Sensor Adaptive Processing
  (CAMSAP)}, 61(6), 2009.

\bibitem{lin2011linearized}
Z.~Lin, R.~Liu, and Z.~Su.
\newblock Linearized alternating direction method with adaptive penalty for
  low-rank representation.
\newblock In {\em Advances in neural information processing systems}, pages
  612--620, 2011.

\bibitem{6854862}
H.~Mansour and A.~Vetro.
\newblock Video background subtraction using semi-supervised robust matrix
  completion.
\newblock In {\em 2014 IEEE International Conference on Acoustics, Speech and
  Signal Processing (ICASSP)}, pages 6528--6532, May 2014.

\bibitem{mu2016scalable}
C.~Mu, Y.~Zhang, J.~Wright, and D.~Goldfarb.
\newblock Scalable robust matrix recovery: Frank--wolfe meets proximal methods.
\newblock {\em SIAM Journal on Scientific Computing}, 38(5):A3291--A3317, 2016.

\bibitem{peng2016connections}
X.~Peng, C.~Lu, Z.~Yi, and H.~Tang.
\newblock Connections between nuclear-norm and frobenius-norm-based
  representations.
\newblock {\em IEEE Transactions on Neural Networks and Learning Systems},
  2016.

\bibitem{peng2016automatic}
X.~Peng, J.~Lu, Z.~Yi, and R.~Yan.
\newblock Automatic subspace learning via principal coefficients embedding.
\newblock {\em IEEE Transactions on Cybernetics}, 2016.

\bibitem{peng2016constructing}
X.~Peng, Z.~Yu, Z.~Yi, and H.~Tang.
\newblock Constructing the l2-graph for robust subspace learning and subspace
  clustering.
\newblock {\em IEEE transactions on cybernetics}, 2016.

\bibitem{shen2014augmented}
Y.~Shen, Z.~Wen, and Y.~Zhang.
\newblock Augmented lagrangian alternating direction method for matrix
  separation based on low-rank factorization.
\newblock {\em Optimization Methods and Software}, 29(2):239--263, 2014.

\bibitem{Sobral2015}
A.~Sobral, T.~Bouwmans, and E.-h. Zahzah.
\newblock Comparison of matrix completion algorithms for background
  initialization in videos.
\newblock In {\em International Conference on Image Analysis and Processing},
  pages 510--518. Springer, 2015.

\bibitem{Sobral20144}
A.~Sobral and A.~Vacavant.
\newblock A comprehensive review of background subtraction algorithms evaluated
  with synthetic and real videos.
\newblock {\em Computer Vision and Image Understanding}, 122:4 -- 21, 2014.

\bibitem{stauffer1999adaptive}
C.~Stauffer and W.~E.~L. Grimson.
\newblock Adaptive background mixture models for real-time tracking.
\newblock In {\em Computer Vision and Pattern Recognition, 1999. IEEE Computer
  Society Conference on.}, volume~2, pages 246--252. IEEE, 1999.

\bibitem{tuzel2005bayesian}
O.~Tuzel, F.~Porikli, and P.~Meer.
\newblock A bayesian approach to background modeling.
\newblock In {\em Computer Vision and Pattern Recognition-Workshops, 2005. CVPR
  Workshops. IEEE Computer Society Conference on}, pages 58--58. IEEE, 2005.

\bibitem{vacavant2012benchmark}
A.~Vacavant, T.~Chateau, A.~Wilhelm, and L.~Lequi{\`e}vre.
\newblock A benchmark dataset for outdoor foreground/background extraction.
\newblock In {\em Asian Conference on Computer Vision}, pages 291--300.
  Springer, 2012.

\bibitem{woo2014robust}
H.~Woo and H.~Park.
\newblock Robust asymmetric nonnegative matrix factorization.
\newblock {\em Computational and Applied Mathematics Reports, University of
  California, USA}, 2014.

\bibitem{wright2009robust}
J.~Wright, A.~Ganesh, S.~Rao, Y.~Peng, and Y.~Ma.
\newblock Robust principal component analysis: Exact recovery of corrupted
  low-rank matrices via convex optimization.
\newblock In {\em Advances in neural information processing systems}, pages
  2080--2088, 2009.

\bibitem{xu2013gosus}
J.~Xu, V.~K. Ithapu, L.~Mukherjee, J.~M. Rehg, and V.~Singh.
\newblock Gosus: Grassmannian online subspace updates with structured-sparsity.
\newblock In {\em Proceedings of the IEEE International Conference on Computer
  Vision}, pages 3376--3383, 2013.

\bibitem{yuan2009sparse}
X.~Yuan and J.~Yang.
\newblock Sparse and low-rank matrix decomposition via alternating direction
  methods.
\newblock {\em preprint}, 12, 2009.

\bibitem{zhou2013moving}
X.~Zhou, C.~Yang, and W.~Yu.
\newblock Moving object detection by detecting contiguous outliers in the
  low-rank representation.
\newblock {\em IEEE Transactions on Pattern Analysis and Machine Intelligence},
  35(3):597--610, 2013.

\bibitem{zivkovic2006efficient}
Z.~Zivkovic and F.~Van Der~Heijden.
\newblock Efficient adaptive density estimation per image pixel for the task of
  background subtraction.
\newblock {\em Pattern recognition letters}, 27(7):773--780, 2006.

\end{thebibliography}

\end{document}